%% file: main.tex
\documentclass[10pt,twocolumn,letterpaper]{article}

\usepackage{iccv}              

\usepackage{multirow}
\usepackage{multicol}

\definecolor{iccvblue}{rgb}{0.21,0.49,0.74}
\usepackage[pagebackref,breaklinks,colorlinks,allcolors=iccvblue]{hyperref}

\def\paperID{7154} 
\def\confName{ICCV}
\def\confYear{2025}

\title{Harmony: A Unified Framework for Modality Incremental Learning}
\author{Yaguang Song\textsuperscript{1},
Xiaoshan Yang\textsuperscript{2,3,1},
Dongmei Jiang\textsuperscript{1},
Yaowei Wang\textsuperscript{1}, 
Changsheng Xu\textsuperscript{2,3,1}\\
\small\textsuperscript{2}Peng Cheng Laboratory, China\\
\small\textsuperscript{1}State Key Laboratory of Multimodal Artificial Intelligence Systems (MAIS), Institute of Automation, Chinese Academy of Sciences (CASIA)\\
\small\textsuperscript{3}School of Artificial Intelligence, University of Chinese Academy of Sciences (UCAS)\\
{\tt\footnotesize songyg01@pcl.ac.cn, xiaoshan.yang@nlpr.ia.ac.cn, 	
jiangdm@pcl.ac.cn, wangyw@pcl.ac.cn, csxu@nlpr.ia.ac.cn}
}
\begin{document}
\maketitle
\input{sec/0_abstract}    
\input{sec/1_intro}
\input{sec/2_related_work}
\input{sec/3_method}
\input{sec/4_exp}
\input{sec/5_conclusion}
{
    \small
    \bibliographystyle{ieeenat_fullname}
    \bibliography{main}
}


\end{document}

%% file: sec/0_abstract.tex
\begin{abstract}
Incremental learning aims to enable models to continuously acquire knowledge from evolving data streams while preserving previously learned capabilities. 
While current research predominantly focuses on unimodal incremental learning and multimodal incremental learning where the modalities are consistent, real-world scenarios often present data from entirely new modalities, posing additional challenges.
This paper investigates the feasibility of developing a unified model capable of incremental learning across continuously evolving modal sequences. 
To this end, we introduce a novel paradigm called \textbf{Modality Incremental Learning (MIL)}, where each learning stage involves data from distinct modalities.
To address this task, we propose a novel framework named \textit{Harmony}, designed to achieve modal alignment and knowledge retention, enabling the model to reduce the modal discrepancy and learn from a sequence of distinct modalities, ultimately completing tasks across multiple modalities within a unified framework.
Our approach introduces the adaptive compatible feature modulation and cumulative modal bridging. 
Through constructing historical modal features and performing modal knowledge accumulation and alignment, the proposed components collaboratively bridge modal differences and maintain knowledge retention, even with solely unimodal data available at each learning phase.
Extensive experiments on the MIL task demonstrate that our proposed method significantly outperforms existing incremental learning methods, validating its effectiveness in MIL scenarios.
The code and datasets will be made publicly available.
\end{abstract}

%% file: sec/1_intro.tex
\section{Introduction}
\label{sec:intro}
With the tremendous advancement of artificial intelligence, deep learning models have demonstrated remarkable performance in various tasks, including visual recognition~\cite{imagenet} and object detection~\cite{objectreview}.
These models are typically trained once and subsequently deployed in practical applications. In real-world scenarios, new data continuously emerges, potentially containing novel categories, which these models struggle to accommodate.
%

\begin{figure}[t]
\centering
    \includegraphics[width=0.9\linewidth]{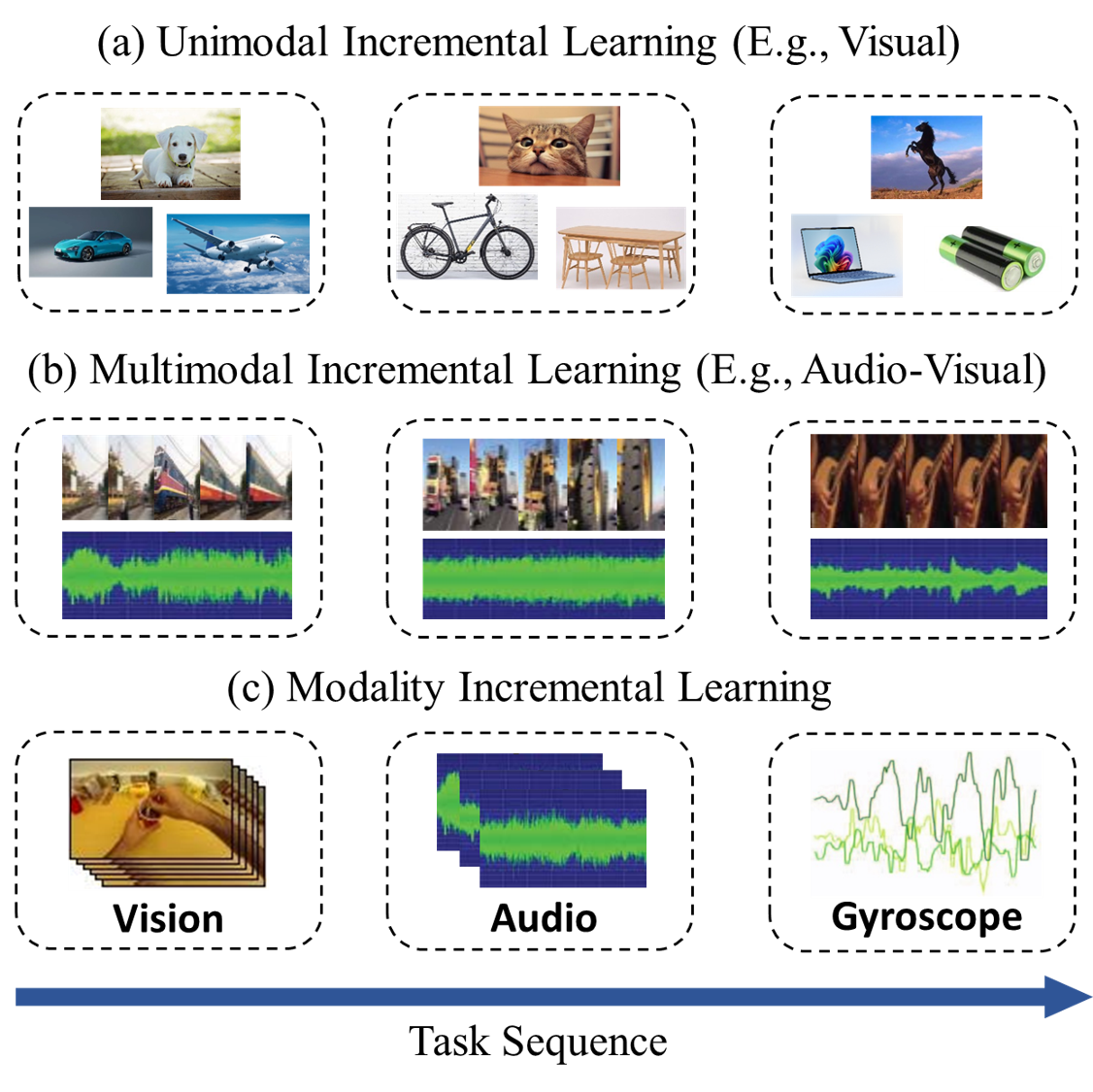}
    \caption{
    Comparison of the (c) Modality Incremental Learning (MIL) with conventional (a) Unimodal Incremental Learning and (b) Multimodal Incremental Learning.
    }
    \vspace{-1.88em}
\label{intro:task111}
\end{figure}

To address this challenge, incremental learning or continual learning methods~\cite{lwf,ewc,wang2024comprehensive} have been proposed. Most incremental learning approaches primarily focus on unimodal scenarios, such as visual classification. Some classical methods preserve knowledge from historical tasks and prevent forgetting by constraining key parameters or utilizing knowledge distillation techniques~\cite{lwf,ewc}. Other approaches retain data from previous tasks and achieve continuous learning through data replay~\cite{icarl, er}, though this approach raises data privacy concerns.
With the ongoing development in the field of incremental learning, these techniques have been extended to multimodal domains. Such methods typically leverage cross-modal correlations to impose constraints on model parameters, thereby facilitating knowledge retention~\cite{vqacl, zuo}.
Although current incremental learning methods have made significant progress, they generally consider scenarios where modalities remain consistent across different stages. Real-world applications, however, often require incorporating new sensory modalities.
%
For instance, a behavior recognition system based on wearable glasses might initially be trained on image data. Over time, the range of modalities available to the device begins to expand, incorporating audio and gyroscope signals. As illustrated in Figure~\ref{intro:task111}, we aim for models to continuously learn capabilities from new modalities while building upon existing competencies, rather than requiring retraining. This approach would significantly reduce the computational and data resources required for model deployment.
Despite its practical importance, the capability to incrementally learn from new modalities remains largely unexplored in the field of incremental learning.

%
In this paper, to investigate whether the model can learn different modalities in a sequential manner, 
we introduce a novel \textbf{Modality Incremental Learning (MIL)} task.
Unlike traditional class incremental learning~\cite{lwf,ewc}, MIL presents a unique scenario where each learning phase involves data from distinct modalities, as illustrated in Figure~\ref{intro:task111}, with access restricted to only current modality data due to privacy concerns~\cite{federated, fetril} or online settings~\cite{mkd}.
We aim to construct a unified model capable of gradually accumulating modal knowledge, both leveraging existing modal knowledge to learn new modalities and preserving knowledge of previous modalities. Some incremental learning methods~\cite{xiao2024configurable, onepeace, xllm, mil} learn different parameters for different tasks, but these approaches cannot address the MIL problem. This limitation stems from their inability to learn associations between different modalities, as well as their requirement for modality identification during inference, which is impractical when only preprocessed features are available.
Compared to traditional unimodal or multimodal incremental learning tasks, MIL faces more severe challenges. 
On one hand, there is the challenge of establishing associations between different modalities and achieving modal alignment when only corresponding modal data can be accessed at different stages. Existing multimodal incremental learning tasks can simultaneously access multiple modalities, thus directly constructing a shared modal space. To better accomplish multimodal tasks, we need to learn the correlations between modalities, which is particularly difficult in this task since we can only access unimodal data at different stages.
On the other hand, there is the challenge of learning new modal knowledge while preserving knowledge of previous modalities, under conditions where data from different stages exhibits substantial modal disparities.
The significant discrepancy between different modalities\cite{mindgap}, coupled with the unavailability of historical data, makes models particularly vulnerable to catastrophic forgetting~\cite{catastrophic}.
Existing incremental learning methods~\cite{lwf, mkd, fineknowledge} often employ distillation based on historical models to maintain knowledge, specifically using historical model outputs to constrain current model outputs based on current task inputs. 
However, these methods cannot effectively address the challenge of MIL.
The reason is that, due to substantial modal disparities, current modal data may not effectively elicit reliable knowledge from historical models for distillation, as these models were trained on fundamentally different modalities.

To address the above challenges of the MIL  task, this paper proposes a novel modal incremental learning framework, named \textit{Harmony}.
The aim of \textit{Harmony} is to achieve modal bridging and unification utilizing only unimodal data at each stage, while simultaneously accumulating modal knowledge to facilitate knowledge retention, enabling the model to learn on a sequence of distinct modalities, ultimately allowing a single model to complete tasks across multiple modalities.
Specifically, we employ a transformer-based~\cite{vaswani2017attention} unified model that has been widely applied in modeling various modalities and has proven its versatility~\cite{vaswani2017attention,visiontransurvey,audiotrans1,audiotrans2}. 
To address the challenges of MIL, our \textit{Harmony} framework incorporates the following synergistic components:
(1) \textbf{Adaptive Compatible Feature Modulation.} 
To achieve modal unification and knowledge retention under conditions where only current modal data is available, we aim to obtain historical modality feature distributions that align with the current modality input.
%
We propose an adaptive feature perturbation construction method that generates augmentation features adaptively based on current modal inputs. By fusing this with historical modal semantic information, we construct more diverse historical modal feature distributions.
(2) \textbf{Cumulative Modal Bridging.} 
Based on the obtained compatible historical modal features, to better utilize historical modalities to assist current modality learning while preserving historical modal knowledge, we introduce the cumulative knowledge aggregation and hybrid alignment to inject historical modal knowledge into the current model. 
Specifically, the proposed cumulative knowledge aggregation first constructs a modality knowledge aggregation module and a gated knowledge adapter. 
We utilize the gated knowledge adapter to filter historical modal knowledge and fuse it with current modal features to achieve the injection of historical modal knowledge at the token level.
Simultaneously, the learned knowledge adapter also preserves historical modal knowledge information. We inject it into the modal knowledge aggregation module while maintaining the overall model structure unchanged.
Through this approach, we continuously inject historical modal knowledge at both feature and parameter levels across different stages, which on one hand assists in learning new modal knowledge, and on the other hand achieves modal knowledge retention.
%
Finally, to map different modalities into a unified space while achieving modal bridging through distillation, we introduce hybrid alignment. 
It achieves modal connection through three complementary strategies: direct feature alignment, contrastive feature alignment, and distribution-level alignment.
The contributions of this work are summarized as follows:
\begin{itemize}
    \item 
    We introduce Modality Incremental Learning (MIL), a novel paradigm designed to explore how models can efficiently extend to new modalities in a continuous manner.
    Our unified framework \textit{Harmony} enables continual learning across distinct modalities, ultimately allowing a single model to complete tasks across multiple modalities.
    \item Our approach achieves effective modal connection and knowledge retention while operating solely with current modality data, via the proposed adaptive compatible feature modulation and cumulative modal bridging.
    \item Extensive experiments demonstrate our approach's superior performance compared to existing incremental learning methods in the MIL setting.
\end{itemize}

%% file: sec/2_related_work.tex
\section{Related Work}
\label{sec:related}

\subsection{Incremental Learning}
The purpose of incremental learning~\cite{wang2024comprehensive, lwf,ewc} is to solve the catastrophic forgetting problem~\cite{catastrophic} of models learning on the evolving data, which refers to the phenomenon of forgetting the knowledge of an old task after learning a new task. 
Researchers have developed several approaches to tackle incremental learning challenges.
Rehearsal-based methods~\cite{icarl, er,diffobject,meng2024diffclass} mitigate forgetting by maintaining a small subset of samples from each learning stage, either through direct storage~\cite{er,icarl} or generative techniques~\cite{diffobject,meng2024diffclass}. 
While these approaches have gained significant traction in research, they face practical limitations due to data privacy~\cite{federated, fineknowledge} or the online environment~\cite{mkd}.
The field has also witnessed the emergence of alternative approaches, including regularization-based methods~\cite{ewc, mas}, architecture-based methods~\cite{pnn, der}, and distillation-based methods~\cite{mkd, multiteacher,lwf}.
In non-exemplar incremental learning scenarios, where historical data access is restricted, distillation-based approaches~\cite{fineknowledge, mkd,li2024fcs} have emerged as the predominant paradigm. 
Zhai et al.~\cite{fineknowledge} innovative fine-grained selective patch-level distillation combined with an old prototype restoration strategy, which effectively balances plasticity and stability. 
Michel et al.~\cite{mkd} advanced the field by exploring momentum knowledge distillation in online incremental learning, achieving state-of-the-art performance.

While incremental learning research has traditionally concentrated on visual recognition tasks, recent years have witnessed the emergence of incremental learning approaches for multimodal tasks~\cite{retrievalcontinual, zhu2023ctp, vqacl, zuo}. 
Wang et al.~\cite{retrievalcontinual} introduced a continual method addressing domain evolution in cross-modal retrieval tasks, primarily utilizing regularization-based methods. 
Zhu et al.~\cite{zhu2023ctp} developed an incremental vision-language pre-training framework, employing topology preservation and momentum contrast to mitigate forgetting in cross-domain image-text pair scenarios.
Recent research has begun to explore whether models can perform incremental learning based on modal sequence tasks~\cite{mil}, where different parameters are learned for different modalities. This is similar to task-incremental settings, which require modality ID during the testing phase. In contrast, the modal incremental learning setup in this paper does not have these requirements, achieving modal incrementality through a unified model with modality knowledge accumulation.

\subsection{Modality Expansion}

Recent years have witnessed remarkable advances in pre-trained models~\cite{gpt, pretrainreview}, with researchers increasingly focusing on developing multimodal pre-trained models~\cite{vlpretrain} that can effectively handle diverse modal tasks. 
Some researchers have attempted to extend existing pre-trained models to new modalities, enabling multimodal capabilities~\cite{audioclip, unival, xllm}.
Notable examples include AudioCLIP~\cite{audioclip}, which aligns CLIP with audio encoders, and UnIVAL~\cite{unival}, which incorporates new modalities through curriculum learning.
Furthermore, Multimodal Large Language Models (MLLMs)~\cite{llava} have shown remarkable progress, primarily through multimodal pre-training that combines LLMs with large-scale multimodal aligned data to enable sophisticated conversational multimodal understanding~\cite{xu2024libra, xllm}.

However, these approaches universally require paired multi-modal data during training. In contrast, our work approaches modal expansion through the lens of incremental learning, where each phase provides data from distinct modalities and the modal identifier is unavailable during inference. This novel paradigm presents unique challenges and fundamentally differentiates our approach from existing methods.


%% file: sec/3_method.tex
\section{Method}
\label{sec:method}

\subsection{Problem Definition}

Modality Incremental Learning (MIL) focuses on developing a unified model with non-stationary data derived from a series of sequential tasks, where each task encompasses data from a distinct modality and shares the same data label space.
The data can be formally represented as \(D=\{D_{t}\}^{T}_{t=1}\), where $T$ denotes the total number of modalities (or phases). 
For modality $t$, \(X_t=\{(x^t_{i}, y^t_{i})\}^{N_t}_{i=1}\) comprises \(N_t\) tuples, including the input data and its corresponding label. 
It is worth noting that when training the model at the $t$-th phase, the model can only access data from the current modality $t$. 
Upon completion of the sequential learning process, the model's performance is evaluated across all previously encountered modalities.

\begin{figure*}[t]
\centering
    \includegraphics[width=0.85\linewidth]{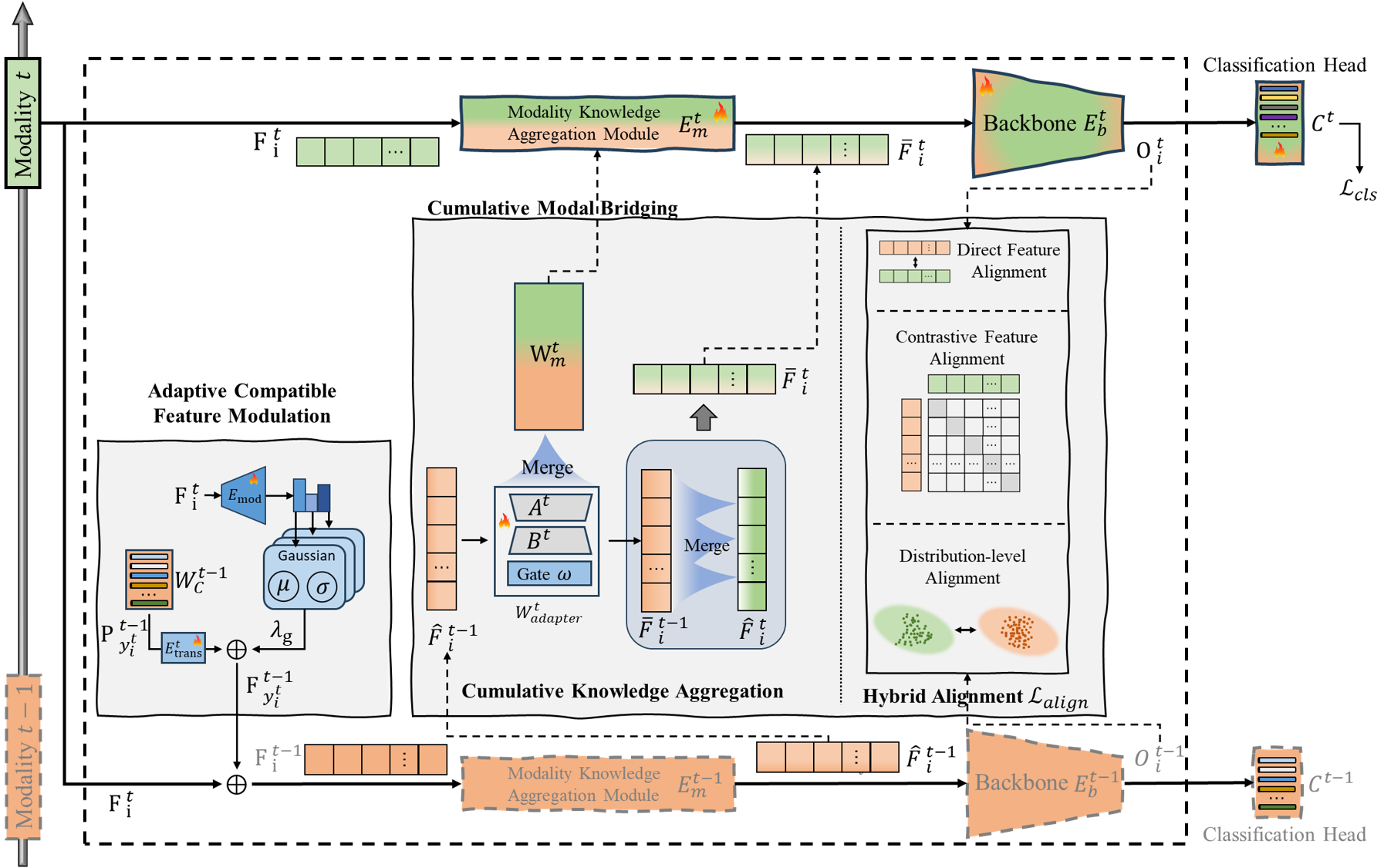}
    \caption{
    Overview of the proposed framework for MIL. At the current phase $t$, we first obtain the modulated feature $F^{t-1}_{i}$ through the adaptive compatible feature modulation for modality $t-1$. Then we integrate the historical modal knowledge through the cumulative knowledge aggregation and achieve modality connection with hybrid alignment $\mathcal{L}_{align}$.
    }
    \vspace{-1.28em}
\label{fig:model}
\end{figure*}

\subsection{Overview}

To overcome the challenges of MIL, we propose a unified framework \textit{Harmony}, which employs a modality bridging paradigm that minimizes inter-modal gaps by simulating compatible historical modal features, thereby facilitating effective knowledge transfer and retention.
The comprehensive framework, illustrated in Figure~\ref{fig:model}, encompasses two key components:
(1) \textbf{Adaptive Compatible Feature Modulation.} 
This component is designed to construct compatible features for the historical model by leveraging current modal features and diversity-enhanced historical modal prototypes. 
We adaptively generate mixed feature perturbations based on current modal inputs to enhance the diversity of historical modal semantic information, thereby simulating historical modal distributions.
%
(2) \textbf{Cumulative Modal Bridging.} 
Based on the constructed historical modal features that match both current modal inputs and historical modal models, we can progressively unify different modal spaces and aggregate diverse modal knowledge to achieve modal knowledge retention.
We first introduce the cumulative knowledge aggregation, including the modality knowledge aggregation module and gated knowledge adapter to achieve modal knowledge accumulation.
We merge the historical modal features filtered by a gated adapter into the intermediate representations of the current modality to preserve historical modal information at the token level. Simultaneously, we merge the learned gating adapter parameters with the modality knowledge aggregation module, facilitating the accumulation of corresponding modal knowledge.
To reduce the discrepancy between modalities and achieve modal unification, we then introduce hybrid alignment, including direct feature alignment, contrastive feature alignment, and distribution level alignment.

\subsection{Architecture}

In this paper, we adopt the transformer architecture~\cite{vaswani2017attention} as the backbone of the modality incremental model, given its proven success and versatility across various modalities~\cite{visiontransurvey,audiotrans1,audiotrans2}. 
Given a raw input $x^{t}_i$ from modality $t$, we first preprocess it to obtain corresponding raw modal features $F^{t}_i \in \mathbb{R}^{L\times d}$, which are then fed into the modality knowledge aggregation module $E^{t}_m$ and backbone network $E^{t}_{b}$ for modeling. $L$ and $d$ are the length and dimension of the feature.
The resulting output features are ultimately passed to the classifier head $C_t$ to complete the classification task.

\subsection{Adaptive Compatible Feature Modulation}

%
Due to the inability to access historical modality data, we can only utilize current modality data. 
This presents us with multiple challenges: on one hand, we cannot directly establish associations between modalities; on the other hand, due to the mismatch between current modal data and historical models, direct knowledge transfer and retention are not feasible.
%
To address the aforementioned challenges and achieve modal bridging, we first utilize current modal features $F^{t}_i$ and prototypes of historical modalities to obtain compatible historical modal features for the historical model through the proposed adaptive compatible feature modulation. 

Given the constraint of data inaccessibility from historical phases, we leverage the weights of the historical classification head as semantic prototypes.
Given the current modal input features \(F^{t}_{i}\), corresponding category labels \(y^{t}_i\), and the weight of historical classification head \(W^{t-1}_{C}\), we first obtain the prototype \(P^{t-1}_{y^{t}_i}\) of the historical modality.
\begin{equation}
  P^{t-1}_{y^{t}_i} = W^{t-1}_{C}[y^{t}_i]
\end{equation}

To obtain more authentic historical modality features and achieve more robust learning, we utilize feature perturbation to enhance prototype diversity, thereby acquiring modality feature distributions with random deviations~\cite{proto}. This approach simultaneously enhances the model's robustness against complex feature variations. 

We propose an adaptive mixture augmentation strategy. The core innovation lies in dynamically generating mixture coefficients conditioned on input features through a lightweight neural network.
Based on the obtained pseudo prototype \(P^{t-1}_{y^{t}_i}\), we construct the modulated feature $F^{t-1}_{y^{t}_i}$ of augmented historical modality with the adaptive mixture augmentation.
\begin{equation} \label{eq:guass}
\begin{aligned}
  F^{t-1}_{y^{t}_i} &= E^{t}_{trans}(P^{t-1}_{y^{t}_i}) + \lambda_{g} \cdot \sum_{i}^{K}\alpha_i * z_i \\
  \alpha_i &= Softmax(E_{mod}(F^{t}_{i}))
\end{aligned}
\end{equation}
where $\lambda_{g}$ represents the overall intensity of feature perturbation, and $z_i \sim \mathcal{N}(0, \sigma_i)$. $E^{t}_{trans}$ is a linear layer for transformation. $K$ represents the number of augmentation distributions. 
We maintain $K$ augmentation components with learnable standard deviations, covering diverse perturbation scales.
A sub-network consisting of two fully-connected layers $E^{t}_{mod}$ takes the current modal features as input, producing normalized mixture coefficients $\{\alpha_i\}_{i=1}^{K}$ via softmax.
By fusing adaptive distributions with different parameters to generate noise based on the semantic characteristics of each input sample, we simulate diverse historical feature distributions. 

Finally, we obtain the modulated feature $F^{t-1}_{i}$ as input for the historical model $M^{t-1}$ by interpolating the $F^{t-1}_{y^{t}_i}$ given current modality input feature \(F^{t}_{i}\).
\begin{equation}
  F^{t-1}_{i} =  F^{t-1}_{y^{t}_i} + F^{t}_{i}
\end{equation}
This feature integrates semantic information from historical modalities, processed through feature perturbation, with the foundation of current modal features.
Through this approach, we construct compatible features $F^{t-1}_{i}$ as the input features for the historical model with only the current modality feature.
Subsequently, we input both the current modal features $F^{t}_{i}$ and the modulated features $F^{t-1}_{i}$ into the unified model. \\
\subsection{Cumulative Modal Bridging} 
Based on the modulated historical modal features and current modal features, to effectively preserve historical modal knowledge while learning current modal knowledge and achieve modal bridging, we propose a cumulative modal bridging method. This approach incorporates a novel cumulative knowledge aggregation technique and hybrid alignment. \\
\textbf{Cumulative Knowledge Aggregation.} We first introduce the cumulative knowledge aggregation, which aims to achieve knowledge retention and modal connection through the fusion of features and parameters across different modalities.
We first employ a modality knowledge aggregation module to establish connections between diverse modalities.
We input both the current modal features and constructed features into two aggregation module $E^{t}_{m}$, $E^{t-1}_{m}$ and obtain $\hat{F}^{t}_{i}$ and $\hat{F}^{t-1}_{i}$, respectively.
\begin{equation}
\begin{aligned}
  \hat{F}^{t}_{i} &= E^{t}_{m}(F^{t}_{i}) \\
  \hat{F}^{t-1}_{i} &= E^{t-1}_{m}(F^{t-1}_{i})
\end{aligned}
\end{equation}
For the acquired intermediate features, we aim to inject historical modal knowledge into the current modality learning process. Specifically, we utilize a gating knowledge adapter to filter the historical modal features.
\begin{equation}
\begin{aligned}
   &\widetilde{F}^{t-1}_{i} = \hat{F}^{t-1}_{i} W^{t}_{adapter} \\
   &W^{t}_{adapter} = \omega \cdot B^{t}A^{t}
\end{aligned}
\end{equation}
The gated knowledge adapter is implemented in a low-rank manner~\cite{hu2021lora}, as shown in Figure~\ref{fig:model}. $W^{t}_{adapter}$ denotes the weight of the low-rank adapter, $A^{t} \in \mathbb{R}^{r \times H}$ and $B^{t} \in \mathbb{R}^{H \times r}$ are the low-rank matrices for computing $W^{t}_{adapter}$, $r$ denotes the rank, $\omega$ denotes the learnable gate.

We subsequently aggregate the transformed historical modal features with the current modal features. Here, we implement token merging through a cross-attention mechanism to obtain the fused features.
\begin{equation}
\begin{aligned}
   &\widetilde{F}^{t}_{i} = Softmax(\frac{\widetilde{F}^{t-1}_{i} {\hat{F}^{t}_{i}}}{\sqrt{d}}) \hat{F}^{t}_{i} \\
\end{aligned}
\end{equation}
where $\widetilde{F}^{t}_{i}$ is the obtained fused feature.

Through this approach, we inject historical modal tokens into the current modality learning process, achieving modal knowledge accumulation at the token level. Simultaneously, we aggregate the learned adapter into the module, accomplishing modal knowledge accumulation while maintaining the model structure unchanged. \\
\textbf{Hybrid Alignment.}\label{hsb} Features $\hat{F}^{t}_{i}$ and $\hat{F}^{t-1}_{i}$ are processed through the transformer backbone network to generate classification features $O^{t}_{i}$ and $O^{t-1}_{i}$.
We introduce hybrid alignment to minimize the discrepancy between the features $O^{t}_{i}$ and $O^{t-1}_{i}$ and achieve knowledge retention.
Given that the input features $O^{t}_{i}$ and $O^{t-1}_{i}$ are not naturally paired data, relying exclusively on direct feature alignment may yield suboptimal modality alignment results. 
We introduce three complementary alignment strategies operating at multiple granularities, including direct feature alignment, contrastive feature alignment, and distribution-level alignment. \\
(1) Direct feature alignment. 
We employ the Mean Square Error (MSE) to enforce direct feature-wise consistency, thereby reducing local discrepancies between corresponding feature representations.
\begin{equation}
\begin{aligned}
   \mathcal{L}^{dir}_{m}(O^{t}_{i}, O^{t-1}_{i}) = \sum_{j}||O^{t}_{i} - O^{t-1}_{i}||_{2}
\end{aligned}
\end{equation}
(2) Contrastive feature alignment. We take the corresponding modulated feature $O^{t-1}_{i}$ as the positive sample for $O^{t}_{i}$ with features of other samples in the batch as negative samples.  
\begin{equation} \label{eq:contra}
\begin{aligned}
   \mathcal{L}^{con}_{m}(O^{t}_{i}, O^{t-1}_{i}) = \sum_{k,j,k\neq j}||{O}^{t}_{k} \cdot {O}^{t-1}_{j} - {O}^{t}_{k} \cdot {O}^{t-1}_{k} + \epsilon||_{2}
\end{aligned}
\end{equation}
where $\mathcal{L}^{con}_{m}(O^{t}_{i}, O^{t-1}_{i})$ enhances discriminative feature learning while maintaining semantic consistency, and $\epsilon$ denotes the margin.\\
(3) Distribution-level alignment. The direct feature alignment and contrastive alignment mainly focus on the individual knowledge of each sample, which might lead to suboptimal alignment. 
Hence, we want to incorporate the underlying knowledge of distributions to better connect different modality spaces.  
Considering that the distributions $D^{t}$ and $D^{t-1}$ are intractable due to the lack of the explicit probability density function, we use the features of batch samples from current modality and modulated pseudo modality data to obtain proxy distribution-level features. 
\begin{equation} 
\begin{aligned}
   &H^{t}({O}^{t}) = \sum_{k=1}\beta_{k}O_{k}, O_{k} \in {O}^{t}, \sum_{k=1}\beta_{k}=1 \\
   &H^{t}(\hat{O}^{t-1}) = \sum_{k=1}\beta_{k}O_{k}, O_{k} \in {O}^{t-1}, \sum_{k=1}\beta_{k}=1
\end{aligned}
\end{equation}
where $H^{t}(O^{t}_{i}, O^{t-1}_{i})$ are the proxy features of distributions $D^{t}$ and $D^{t-1}$, $O^{t}_{i}$ and $O^{t-1}_{i}$ are the features of batch samples from current modality and modulated features. $\beta_{k} \in [0, 1]$ denotes the weight of the $k$-th features, which is learnable. The distribution-level alignment is conducted as follows: 
\begin{equation}
\begin{aligned}
   \mathcal{L}^{dis}_{m}(H^{t}({O}^{t}), H^{t}({O}^{t-1})) = ||H^{t}(O^{t}) - H^{t}(O^{t-1})||_{2}
\end{aligned}
\end{equation}
The overall alignment loss is denoted as $\mathcal{L}_{align}$.
\begin{equation}\label{eq:align}
\begin{aligned}
   \mathcal{L}_{align} = \mathcal{L}^{dir}_{m}(O^{t}_{i}, O^{t-1}_{i}) + \lambda_{con}\cdot\mathcal{L}^{con}_{m}(O^{t}_{i}, O^{t-1}_{i}) + \\ \lambda_{dis}\cdot\mathcal{L}^{dis}_{m}(H^{t}(O^{t}), H^{t}(O^{t-1}))
\end{aligned}
\end{equation}
where $\lambda_{con}$ and $\lambda_{dis}$ serve as trade-off weights for balancing different losses. With hybrid alignment, we can reduce gaps between different modalities and achieve modal connection.\\

\subsection{Overall Objective}

The overall objective function of the proposed method for MIL combines the alignment loss, and classification loss:
\begin{equation} \label{eq:loss}
\begin{aligned}
    \mathcal{L}_{all} = \mathcal{L}_{cls} + \lambda \cdot \mathcal{L}_{align}
\end{aligned}
\end{equation}
where $\mathcal{L}_{cls} = \frac{1}{N_t}\sum_{k=1}^{N_t}l_{ce}(C^{t}(O^{t}_{k}), y)$ represents the supervised classification loss of the current modality task, $l_{ce}$ denotes the Cross-entropy loss, $\mathcal{L}_{align}$ denotes the modality alignment loss, $\lambda$ serves as the trade-off weight for balancing different losses.

%% file: sec/4_exp.tex
\section{Experiments}
\label{sec:exp}

\subsection{Modality Incremental Learning benchmark}
Given the absence of dedicated modal incremental learning datasets, we establish challenging MIL benchmarks by adapting existing multimodal datasets for comprehensive model evaluation.\\
\textbf{EPIC-MIL} 
We derive our MIL dataset from Epic-Kitchens-100~\cite{epic100}, the largest multimodal dataset in egocentric Human Activity Recognition (HAR).
For the MIL task, we structure the dataset into three distinct phases encompassing RGB, Flow, and Audio modalities.
The dataset maintains a unified category space across all modalities, utilizing 93 verb labels for action classification.
Each modality encompasses 21,117 samples, split into training (70\%), val (10\%), and test sets (20\%) to facilitate evaluation. \\
\textbf{Drive\&Act-MIL} We derive our Drive\&Act-MIL dataset from the Drive\&Act~\cite{martin2019drive} dataset, which is a multimodal benchmark for fine-grained categorization of driver behavior. 
For the MIL task, we structure the dataset into three distinct phases encompassing RGB, depth, and infrared (IR) modalities.
The dataset maintains a unified category space across all modalities, utilizing 39 semantic activities for drive behavior categorization.
Each modality encompasses 10,017 samples, split into training (70\%), val (10\%), and test sets (20\%) to facilitate evaluation.

\subsection{Metrics}
We formulate the metrics from two aspects to evaluate the performance of different methods on the MIL task. 
First, we adopt the Average Accuracy from incremental learning~\cite{wang2024comprehensive} to evaluate the overall performance of the model on learned modalities, balancing plasticity and stability. 
Specifically, $S_{m,n}$ represents the classification results evaluated on the $n$-th modality after incremental learning of the $m$-th modality, and the average accuracy $AA_{m}$ can be expressed as:
\begin{equation}
\begin{aligned}
   AA_{m} = \frac{1}{V_m}\sum_{n=1}^{V_m}S_{m, n}
\end{aligned}
\end{equation}
where $V_m$ denotes the number of modalities. 
Furthermore, unlike traditional class incremental learning, MIL distinctively handles an expanding set of modalities.
Therefore, we conduct multimodal classification evaluation on the final model obtained after learning across all modalities, using the multimodal classification accuracy $A_{multi}$ as a metric to evaluate the model's capability in multimodal representation modeling. 
We adopt a late fusion approach for multimodal classification. Different modalities of the same sample are input into the learned model, and their outputs are averaged as the final output for classification evaluation.

\subsection{Implementation Details}
We adopt the transformer (Base) model~\cite{vaswani2017attention} as our backbone model, which is trained from scratch in the experiment.
For the EPIC-MIL and Drive\&Act-MIL datasets, raw features are extracted and fed into models. For further details, please refer to the supplementary materials.
We adopt a linear layer for the modality knowledge aggregation module. The gated knowledge adapter is implemented in a low rank manner (rank 128)~\cite{hu2021lora}.
At each phase, models are trained for 50 epochs on 2 NVIDIA A100 GPUs with batch size 256 per GPU. 
We use the AdamW optimizer with a weight decay of 0.05, and the learning rate is set as 5e-4. 
The parameter $\lambda_{g}$ for the adaptive modulation in Eq.~\ref{eq:guass} is set to 0.6. 
The number $K$ of feature perturbations in Eq.~\ref{eq:guass} is set to 3.
The margin factor $\epsilon$ in Eq.~\ref{eq:contra} is set to 0.3. The weights $\lambda_{con}$ and $\lambda_{dis}$ in Eq.~\ref{eq:align}
are set to 0.8 and 0.6. The trade-off weight $\lambda$ for losses  in Eq.~\ref{eq:loss} is set to 1.5.

\subsection{Comparison Methods}
To verify the effectiveness of our method, we conduct comprehensive comparisons against representative incremental learning approaches. 
Joint training (\textbf{JointT}) with all available samples serves as the performance upper bound, while Sequential Learning (\textbf{SeqL}), which processes different modalities sequentially without additional operations, establishes the baseline.
We adopt several basic methods, such as the full parameter regularization (\textbf{FullR}), which uses $l_2$ loss to maintain consistency between current and historical models.
\textbf{Frozen} represents maintaining a fixed backbone network while allowing only classifier parameter updates.
Additionally, we include the classic regularization-based method \textbf{EwC}~\cite{ewc}, the classic distillation-based method \textbf{LwF}~\cite{lwf}. 
We evaluate against two representative State-of-the-Art distillation-based methods \textbf{FGKSR}~\cite{fineknowledge} and \textbf{MKD}~\cite{mkd}. 


\begin{table}[t]
    \setlength{\tabcolsep}{3pt}{
    \begin{tabular}{c|cc|cc}
    \hline
    \multirow{2}{*}{Method}     &  \multicolumn{2}{c|}{EPIC-MIL}  &  \multicolumn{2}{c}{Drive\&Act-MIL}  \\ 
         & $AA_3(\uparrow)$ &  $A_{multi}(\uparrow)$ & $AA_3(\uparrow)$ & $A_{multi}(\uparrow)$  \\
    \hline
    SeqF     &  29.81&    34.35&  38.42 
&   67.12
\\
    FullR     &  29.98&    39.20&  36.94 
&    55.84
\\
    Frozen     &  27.93&    33.67&  40.50 
&   66.92
\\
    EwC~\cite{ewc}     &  29.84&    34.15&  39.65 
&   68.55
\\
    LwF~\cite{lwf}     &  31.94&    39.46&  39.44 
&   59.72
\\
    FGKSR~\cite{fineknowledge}     &  32.31&    37.61&  50.52 
&   69.47
\\
    MKD~\cite{mkd}     &  33.78&    45.36&  49.93 
&   68.58
\\
    \textbf{Ours}     &  \textbf{37.48}&   \textbf{50.30}&  \textbf{55.07} 
&   \textbf{75.85}
\\
    \hline
    JointT    &  44.84&   62.22&  80.04 &   86.23\\
    \hline
    \end{tabular}}
    \caption{Experimental Results on EPIC-MIL and Drive\&Act-MIL dataset. The best scores are \textbf{BOLD}.}
    \label{tab:exp}
    \vspace{-1.88em}
\end{table}

\begin{table*}[t]
    \centering
    \setlength{\tabcolsep}{11pt}{
    \begin{tabular}{c|c|cc|ccc|cc}
    \hline
    \multirow{2}{*}{Method}     & RGB(Acc)&  \multicolumn{2}{c|}{Flow (Acc)}  &  \multicolumn{3}{c|}{Audio (Acc)}  & \multirow{2}{*}{$AA_3(\uparrow)$} &  \multirow{2}{*}{$A_{multi}(\uparrow)$} \\ \cline{2-7}
         & $S_{1,1}$ & $S_{2,2}$ & $S_{2,1}$ & $S_{3,3}$ & $S_{3,2}$ & $S_{3,1}$ &  &   \\
    \hline
    w/o ACFM&  44.26&  52.74&  33.25&  44.50&  35.15&  24.73&  34.79&   44.68\\
    w/o AFP&  44.26&  52.20&  33.65&  44.67&  35.66&  25.43&  35.26&   45.56\\
    w/o CMB&  44.78&  52.28&  32.91&  43.36&  32.35&  22.07&  32.59&   41.36\\
    w/o CKA&  44.26&  51.06&  31.41&  42.49&  33.19&  23.73&  33.13&   44.52\\
    w/o $\mathcal{L}_{align}$&  44.26&  52.70&  33.20&  44.50&  33.14&  23.71&  33.78&   44.60\\
    w/o DirAlign&  44.26&  52.38&  32.71&  43.72&  35.01&  25.09&  34.61&   46.58\\
    w/o ContAlign&  44.26&  52.50&  32.75&  43.70&  36.58&  26.13&  35.47&   47.30\\
    w/o DistAlign&  44.26&  52.62&  32.73&  43.96&  35.97&  25.37&  35.10&   46.76\\
    \hline
     \textbf{Ours}     &  44.26&  52.48&  32.75&  43.70&  41.60&  27.13&  \textbf{37.48}&   \textbf{50.30}\\
    \hline
    \end{tabular}}
    \caption{Ablation study of key components of our method on EPIC-MIL dataset.}
    \label{tab:ablation}
\end{table*}

\begin{table*}[t]
    \centering
    \setlength{\tabcolsep}{11pt}{
    \begin{tabular}{c|c|cc|ccc|cc}
    \hline
    \multirow{2}{*}{Method}     & Audio (Acc) &  \multicolumn{2}{c|}{RGB (Acc)}  &  \multicolumn{3}{c|}{Flow (Acc)}  & \multirow{2}{*}{$AA_3(\uparrow)$} &  \multirow{2}{*}{$A_{multi}(\uparrow)$} \\ \cline{2-7}
         & $S_{1,1}$ & $S_{2,2}$ & $S_{2,1}$ & $S_{3,3}$ & $S_{3,2}$ & $S_{3,1}$ &  &   \\
    \hline
    SeqF     &  43.52&  43.93&  22.13&  50.46&  21.33&  12.02&  27.94&   42.60\\
    FullR     &  43.52&  26.73&  39.78&  43.52&  19.87&  38.54&  33.98&   43.26\\
    Frozen     &  43.52&  38.70&  12.64&  48.48&  22.97&  8.94&  26.80&   39.30\\
    EwC~\cite{ewc}     &  43.52&  43.86&  22.56&  50.34&  21.56&  13.14&  28.35&   42.61\\
    LwF~\cite{lwf}&  43.52&  42.64&  28.31&  43.52&  32.73&  23.43&  31.89&   44.10\\
    FGKSR~\cite{fineknowledge}     &  43.52&  42.73&  19.21&  51.96&  32.83&  15.93&  33.57&   44.86\\
    MKD~\cite{mkd}     &  43.52&  38.09&  35.37&  42.09&  27.06&  35.79&  34.98&   39.95\\
    \textbf{Ours}     &  43.19&  43.35&  31.06&  50.70&  33.85&  26.97&   \textbf{37.17}&   \textbf{49.62}\\
    \hline
    JointT     &  -&  -&  -&  49.34&  41.64&   43.53&  44.84&   62.22 \\
    \hline
    \end{tabular}}
    \caption{Experimental Results on EPIC-MIL dataset with modality order Audio-RGB-Flow.}
    \label{tab:order}
    \vspace{-1.88em}
\end{table*}

\subsection{Experimental Results}
Table~\ref{tab:exp} presents comprehensive experimental results on the EPIC-MIL (RGB(1)-Flow(2)-Audio(3)) and Drive\&Act-MIL datasets (RGB(1)-Depth(2)-IR(3)). 
The evaluation metrics encompass average accuracy metrics, and multimodal classification metrics $A_{multi}$.\\
\textbf{Average Accuracy $AA_{3}$. }
First, we focus on the average accuracy metric, which represents the model's comprehensive performance in incremental learning tasks. 
Our proposed approach achieved state-of-the-art performance on both datasets, surpassing the SOTA baseline by 3.7\% and 4.5\%. These results validate the effectiveness of our \textit{Harmony} framework in simultaneously accommodating new modality learning while preserving knowledge from previous modalities.
Among basic and classical incremental learning methods, LwF demonstrated superior performance, validating the efficacy of distillation based approaches in MIL scenarios. 
MKD emerged as a strong baseline method, leveraging momentum-based historical modality model updates to enhance adaptability to current modality inputs. This performance highlights the crucial impact of modality discrepancies and model-input mismatches on distillation effectiveness.
The FGKSR method, despite employing historical modality prototypes for constraints and classification, achieved suboptimal results compared with Ours. This suggests that direct prototype-based constraints prove insufficient for knowledge preservation in MIL due to substantial inter-modal differences.
\\
\textbf{Multimodal classification $A_{multi}$. } To better evaluate the multimodal modeling capability of each model after sequential modal learning, we conduct multimodal classification using the final trained models. 
This evaluation specifically examines classification accuracy when integrating predictions across multiple modalities.
The multimodal classification results $A_{multi}$ presented in Table~\ref{tab:exp} demonstrate that our approach achieves superior performance with 50.3\% accuracy and 75.85\%, substantially surpassing comparative methods.
This exceptional performance in multimodal classification validates two crucial aspects of our approach: first, its effectiveness in maintaining cross-modal knowledge preservation, and second, its superior multimodal representation learning through the \textit{Harmony} framework.

\subsection{Ablation Study}
We conduct comprehensive ablation studies to validate the effectiveness of each key component in our proposed method, with results presented in Table~\ref{tab:ablation}.
\\
\textbf{Adaptive Compatible Feature Modulation. } w/o ACFM denotes the model where we remove the adaptive compatible feature modulation and simply input the feature of the current modality to the historical model. We can see from the results that Ours outperforms w/o ACFM, which validates the importance of the compatibility of the input feature and historical models for modality connection and knowledge retention. 
We further validated the effectiveness of our adaptive mixed feature perturbation design. w/o AFP represents the approach of directly using a single noise augmentation to obtain compatible historical features, which shows decreased performance compared to Ours, thus demonstrating the effectiveness of adaptive feature perturbation.
\\
\textbf{Cumulative Modal Bridging. } 
The Cumulative Modal Bridging component demonstrates a significant impact on model performance. When it is removed (w/o CMB), performance notably deteriorates. This decline confirms the proposed component's crucial role in minimizing inter-modal gaps and enhancing knowledge transfer and retention across modalities.
\\
\textbf{Cumulative Knowledge Aggregation. } 
Cumulative Knowledge Aggregation (CKA) exhibits a substantial contribution to model effectiveness. Removing the modal knowledge aggregation process leads to performance degradation. This validates the significance of our approach in aggregating different modal knowledge and achieving modal bridging.
\\
\textbf{Hybrid Alignment. } 
Hybrid Alignment emerges as a vital component for modal alignment. 
The model variant without the hybrid semantic alignment loss $\mathcal{L}_{align}$ (w/o $\mathcal{L}_{align}$), which lacks explicit regularization, demonstrates inferior performance. 
This confirms the component's effectiveness in reducing cross-modal discrepancies and enhancing overall model performance.
We also conducted experiments by individually removing the three alignment constraints from our framework (w/o DirAlign, w/o ContAlign, and w/o DistAlign). The results show performance degradation compared to our complete method across all variants, demonstrating the effectiveness of these constraints in facilitating modality bridging.

\begin{figure}[t]
\centering
    \includegraphics[width=0.9\linewidth]{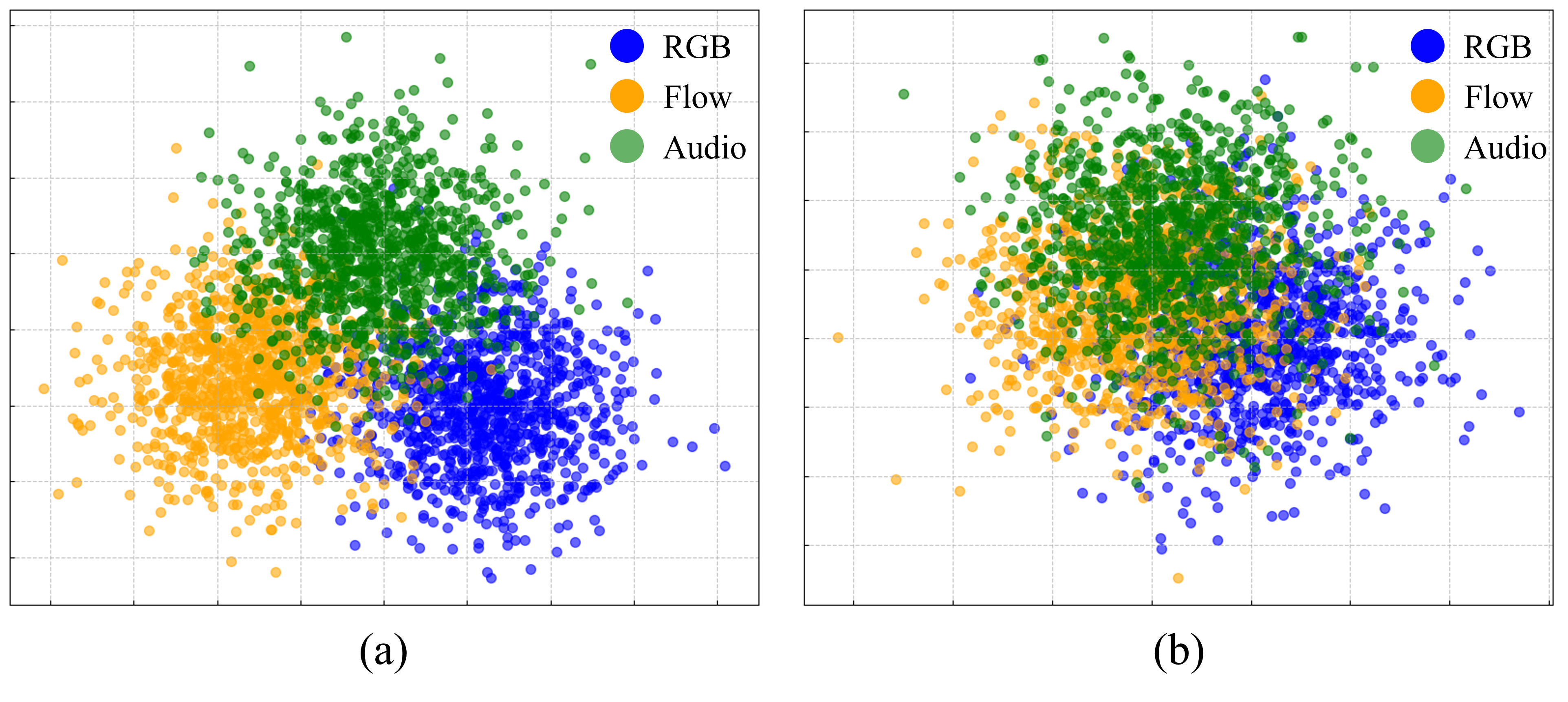}
    \caption{
    (a) Raw input features $F_{i}^{t}$ of RGB, Flow, and Audio. (b) Features $\hat{F}_{i}^{t}$ processed by the modality knowledge aggregation module.
    }
\label{visualization}
\end{figure}

\begin{figure}[t]
\centering
    \includegraphics[width=1\linewidth]{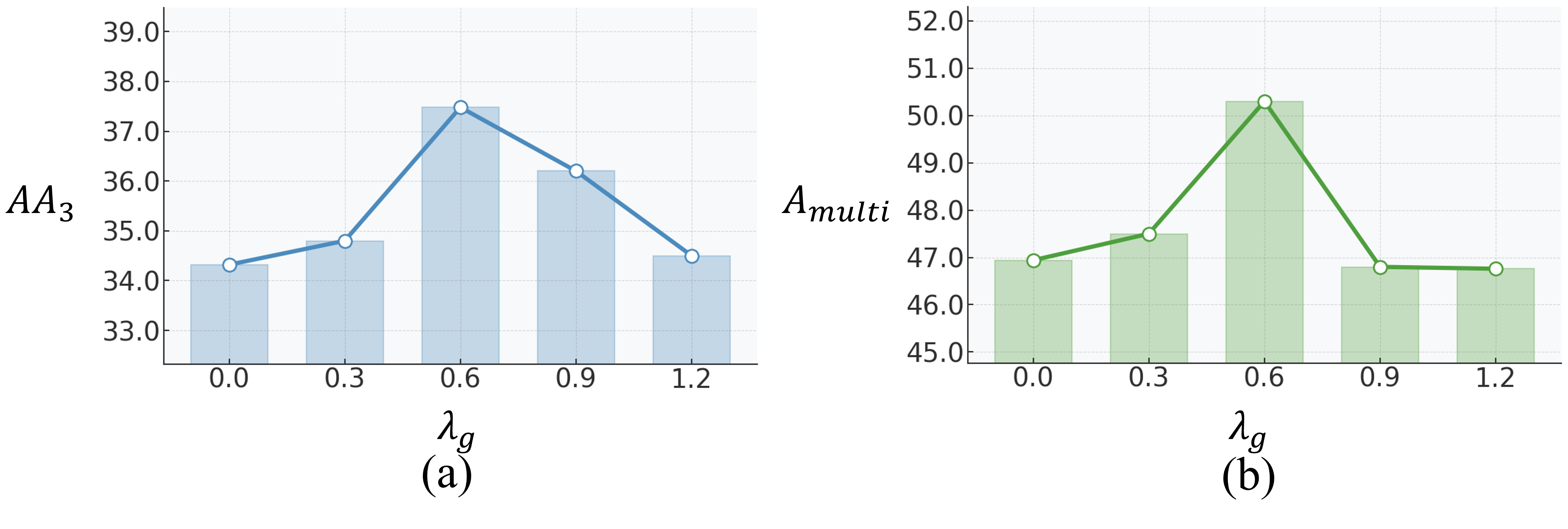}
    \caption{
    Plot of analysis of hyperparameter $\lambda_{g}$, which is the intensity of feature perturbation for the feature modulation. 
    }
\label{hyper}
\end{figure}

\begin{figure}[t]
\centering
    \includegraphics[width=1\linewidth]{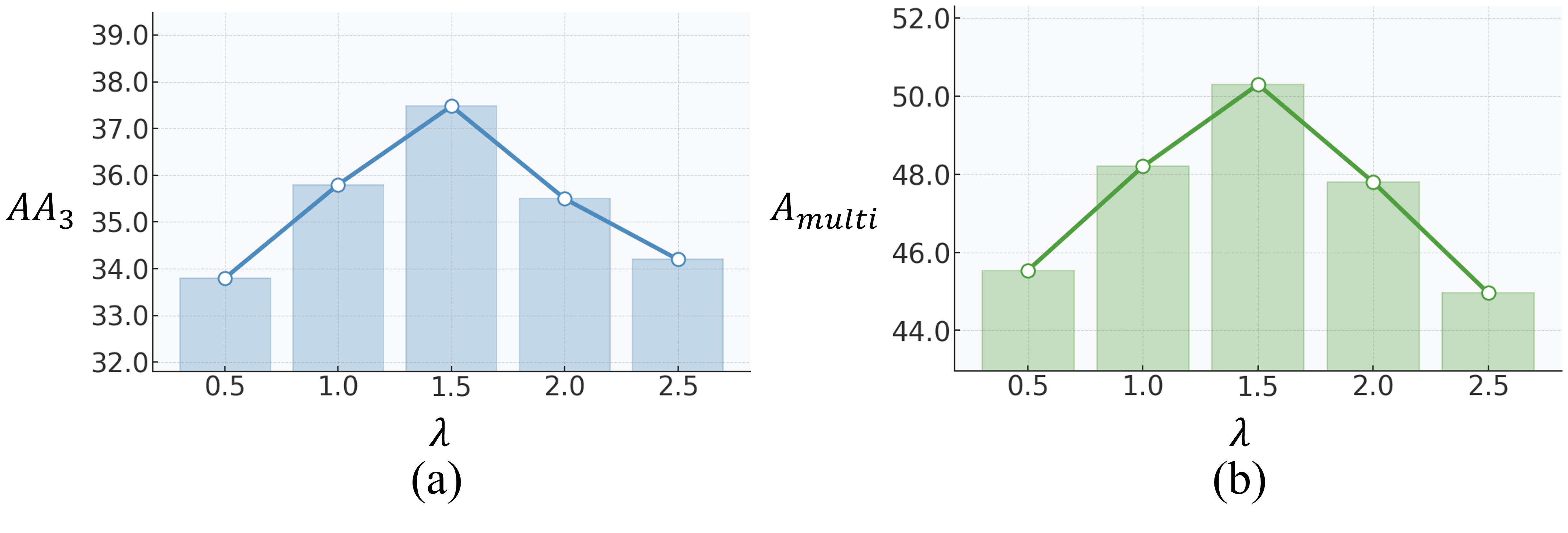}
    \caption{
    Plot of analysis of hyperparameter $\lambda$, which is a trade-off weight for balancing different losses. 
    }
\label{hyper2}
\end{figure}

\subsection{Further Remarks}
\textbf{Influence of Modality Order. } 
Given the inherent differences between modalities and their varying degrees of distributional gaps, the sequential order of modal learning potentially influences model performance. To systematically investigate this impact in MIL, we conducted comprehensive experiments on the EPIC-MIL dataset by altering the modal sequence from RGB-Flow-Audio to Audio-RGB-Flow.
The comparative results presented in Table~\ref{tab:order} reveal consistent performance patterns across different modal sequences. 
Overall performance metrics remain relatively stable across all evaluated methods when compared to the original sequence, suggesting minimal sensitivity to modal ordering. 
Some models are influenced by modal sequences, leading to a certain degree of variation in performance at different stages.
Notably, our proposed approach maintains its superior performance across different sequential configurations, consistently outperforming baseline methods. 
This robust performance across varying modal sequences demonstrates not only the effectiveness of our method but also its stability and adaptability in handling different modal learning sequences.
\\
\textbf{Visualization.} 
To provide intuitive insights into our method's effectiveness, we present a comprehensive visualization analysis. Figure~\ref{visualization} illustrates feature distributions using t-SNE~\cite{tsne}, with samples randomly selected from the EPIC-MIL test dataset sharing identical class labels.
Specifically, Figure~\ref{visualization}(a) depicts the distribution of raw input features across RGB, Flow, and Audio modalities, while Figure~\ref{visualization}(b) showcases the corresponding features after processing through our modality knowledge aggregation module. 
The comparative visualization reveals a substantial reduction in inter-modal discrepancies.
This visual evidence substantiates the effectiveness of our proposed modality bridging mechanism in harmonizing multi-modal representations while preserving semantic consistency.
\\
\textbf{Analysis of Hyperparameter $\lambda_{g}$.}
We adopt feature perturbation in cross-modal feature modulation for compatible historical modal features. 
The intensity parameter $\lambda_{g}$ in Eq.~\ref{eq:guass} plays a critical role as it governs the overall magnitude of deviation from the historical prototypes. 
As observed in Figure~\ref{hyper}, the overall performance of our method first increases and then decreases as this parameter increases, achieving optimal results at 0.6.
This pattern indicates an optimal trade-off point where Gaussian augmentation enhances the diversity of features and model generalization through appropriate feature perturbation while avoiding excessive noise that could compromise performance.
The empirical results validate our approach's ability to leverage controlled stochastic augmentation for improved cross-modal feature adaptation.
\\
\textbf{Analysis of Hyperparameter $\lambda$.}
In the proposed \textit{Harmony} method, the hyper-parameters $\lambda$ is the balance weights of the loss function.
Figure~\ref{hyper2} shows the impact on the EPIC-MIL dataset.
As shown, our model achieves the best performance when $\lambda$ is set to $1.0$.
The performance decreases as $\lambda$ approaches 0, which demonstrates the necessity of the hybrid alignment in modality alignment and knowledge retention. We also observe a performance drop when using larger values of $\lambda$ because they may impact the integration of current modality knowledge.

%% file: sec/5_conclusion.tex
\section{Conclusions}
\label{sec:conclusion}

In this paper, we introduce a modality incremental learning paradigm to investigate whether models can continuously learn new modalities. To address the unique challenges, we propose the \textit{Harmony} framework, which enables continual learning across sequential distinct modalities. 
The cornerstone of our approach lies in achieving effective modal connection and cumulative knowledge integration with the adaptive compatible feature modulation and cumulative modal bridging, utilizing solely unimodal data from each stage.
To validate the effectiveness of our approach, we have constructed two modality incremental learning datasets, EPIC-MIL and Drive\&Act-MIL.
Through the comprehensive empirical evaluation of the proposed dataset, our method consistently outperforms competitive baseline approaches across various performance metrics. The experimental results validate both the effectiveness and superiority of our framework under different modal learning sequences.
In the future, we plan to extend our work to encompass additional modalities and explore modality incremental learning challenges in open-world scenarios.